\documentclass[letterpaper, 10 pt, conference]{ieeeconf}  

\IEEEoverridecommandlockouts                              
\usepackage[pdftex]{graphicx}
\usepackage{tabularx}
\usepackage{multirow}

\usepackage{amsmath}
\usepackage{amssymb}
\usepackage{amsfonts}
\usepackage[ruled, vlined, linesnumbered]{algorithm2e}

\usepackage{booktabs}

\usepackage{balance}

\usepackage{graphicx, subcaption}

\usepackage{url}

\usepackage{xcolor}

\usepackage[font=small]{caption}

\usepackage{balance}

\usepackage{tikz}
\usetikzlibrary{automata, positioning, arrows}

\usepackage{todonotes}

\title{Path Planning with Potential Field-Based Obstacle Avoidance in a 3D Environment by an Unmanned Aerial Vehicle}

\author{Ana Batinovic, Jurica Goricanec, Lovro Markovic, Stjepan Bogdan
\thanks{Authors are with the University of Zagreb, Faculty of Electrical Engineering  and Computing, LARICS Laboratory for Robotics and Intelligent Control Systems, Unska 3, 10000 Zagreb, Croatia; {\tt\small (ana.batinovic, jurica.goricanec, lovro.markovic, stjepan.bogdan)@fer.hr}}}

\begin{document}

\maketitle

\begin{abstract}
In this paper we address the problem of path planning in an unknown environment with an aerial robot.
The main goal is to safely follow the planned trajectory by avoiding obstacles. The proposed approach is suitable for aerial vehicles equipped with 3D sensors, such as LiDARs. It performs obstacle avoidance in real time and on an on-board computer. We present a novel algorithm based on the conventional Artificial Potential Field (APF) that corrects the planned trajectory to avoid obstacles. To this end, our modified algorithm uses a rotation-based component to avoid local minima. The smooth trajectory following, achieved with the MPC tracker, allows us to quickly change and re-plan the UAV trajectory.
Comparative experiments in simulation have shown that our approach solves local minima problems in trajectory planning and generates more efficient paths to avoid potential collisions with static obstacles compared to the original APF method.

\end{abstract}

\IEEEpeerreviewmaketitle

\section{Introduction}

Unmanned aerial vehicles (UAVs) have been recently utilized in various applications, such as agriculture \cite{Tsouros2019}, wind turbine inspection \cite{Car2020}, inspection of civil infrastructure \cite{Greenwood2019}, search and rescue \cite{Shakhatreh2019}, autonomous exploration \cite{Batinovic-RAL-2021}, etc. To successfully accomplish their missions, UAVs need to execute safe and collision-free paths. 
Usually, the information about the obstacle is not known beforehand and should be detected by the sensors attached to the UAV during the mission. The UAV must be able to avoid the detected obstacle using the information collected online. In other words, obstacle detection and avoidance is one of the most important capabilities for successful UAV application execution.

There are a variety of algorithms to achieve safe path planning \cite{Radmanesh2018}. One of the most attractive methods is the APF method \cite{Khatib1986}. The principle by which the APF method navigates the UAV through the environment is the generation of a virtual force vector that dictates the direction of UAV motion. This resulting virtual force vector is the sum of all attractive and repulsive forces acting on the UAV. Attractive forces pull the UAV towards the unknown environment, while repulsive forces push it away from obstacles. The APF method offers efficient and effective way for robot obstacle avoidance. 

\begin{figure}[t!]
	\centering
	\includegraphics[width=0.95\columnwidth]{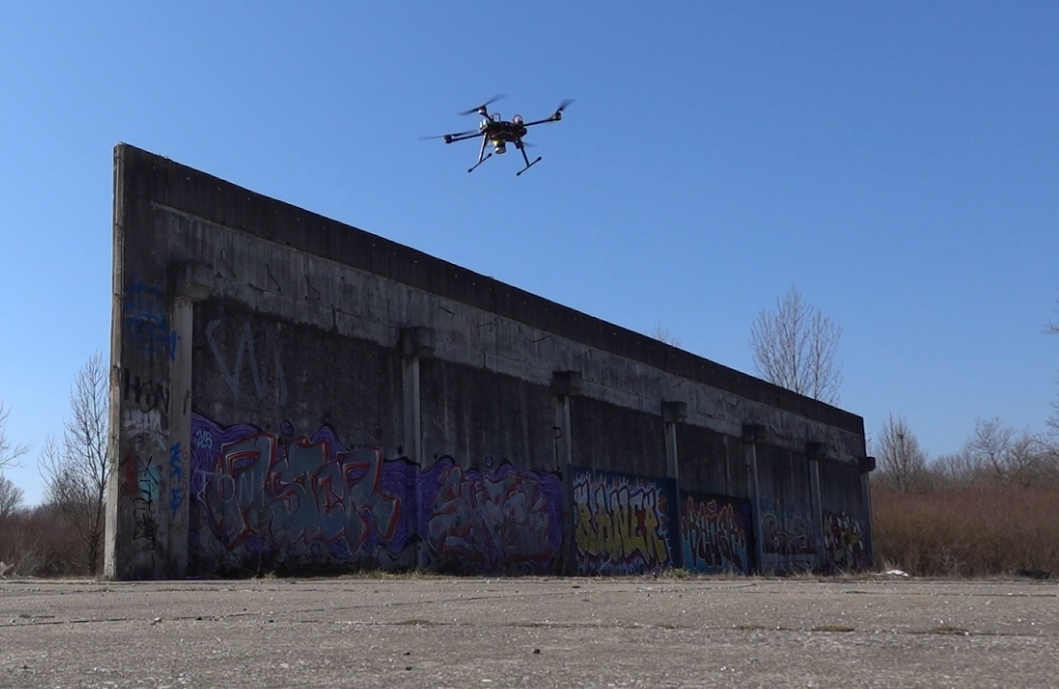}
	\caption{A real-world experiment. The UAV successfully avoids an obstacle.}
	\label{fig:experiment}
\end{figure}

However, the APF method has several major limitations \cite{Choi2020}, the most common of which is local minima. 
To overcome the local minima problem, we propose a potential field-based algorithm that follows the planned trajectory while making corrections to the trajectory when repulsive forces are active (near obstacles). While the attractive potential field is zero, the repulsive potential field is composed of the translational and rotational components of the repulsive force. The algorithm results in fully autonomous path planning with obstacle avoidance as shown in Fig. \ref{fig:experiment}. The solution includes on-board trajectory planning and obstacle avoidance.
The contributions of this paper are summarized as follows:
\begin{itemize}
    \item The modified potential field algorithm based on translational and rotational repulsive force.
    \item Resolution of local minima using the rotational component of the repulsive force.
    \item Integration of the MPC tracker to achieve smooth trajectory following.
    \item Comprehensive analysis and validation of the proposed approach in both simulation and real-world experiments.
\end{itemize}

In Section \ref{sec:related} we give an overview of the state-of-the-art in obstacle avoidance methods and position our work in relation to them.  Section \ref{sec:proposed} is the core of the paper and contains the detailed explanation of the proposed method. The results of the simulations and experiments performed with a UAV and their analysis are presented in Sections \ref{sec:simulation} and \ref{sec:experiment}. The paper ends with a conclusion in Section \ref{sec:conclusion}.

\section{Related work}
\label{sec:related}

The potential field method is commonly used in path planning because it has many advantages, such as simple implementation, simple structure and low computational complexity. The method is based on the concept of a repulsive and attractive force field to repel a robot from an obstacle and attract it toward the target \cite{Radmanesh2018}, \cite{Khatib1986}, \cite{Yasin2020}. 

Even though the basic approach to obstacle avoidance developed in \cite{Khatib1986} is efficient and simple, the path plan may not be globally optimal, especially if the obstacles are close to the robot. In literature, there has been a significant amount of improved approaches based on the APF method applied to path planning of aerial robots.
Firstly, integrating other algorithms into the APF method, such as Rapidly-exploring Random Tree \cite{Gao2018}, \cite{Xinyu2019}, Particle Swarm Optimization \cite{Ahmed2015}, etc. Secondly, modifying the original APF method, e.g., by extending the repulsion formula to include more terms \cite{Sun2017} or by using a different form of the potential function, as described in \cite{Weerakoon2015}, \cite{Sudhakara2018}. 
In \cite{Mac2016}, the authors enumerate the problems of the traditional potential field method and focus on problems with unreachable targets. The Modified Potential Field Method (MPFM) compensates for the repulsive force by adding the Euclidean distance that connects the attractive force to the repulsive force.  The authors point out that the potential field method has several inherent limitations in which the non-reachable target problem is the most challenging.
Zhu et al. \cite{Zhu2016} proposed a Modified Artificial Potential Field algorithm (MAPF), which is able to decompose the total force and estimate the physical barriers on the 3D environment.

Another approach of work on the APF method utilizes a global path planning method to find a desired path to the goal and uses attractive and repulsive concepts of the APF methods for local path planning \cite{Chiang2015}, \cite{Patil2011}, \cite{Jaillet2010}.  Jaillet et al. \cite{Jaillet2010} used a user-defined cost-map to influence node placement in an RRT algorithm. The cost-map specifies a repulsion or attraction factor for each region. Similarly, in \cite{Patil2011} Navigation Fields assign a gradient that a robot follows. In the same manner, Scherer et al. \cite{Scherer2008} combine global and local potential field-based planners to navigate an UAV toward the goal. The global planner is based on the implementation of a Laplace equation that generates a potential function with an unique minimum at the target, while the local planner uses a modification of the conventional potential field method in which the relative angles between the goal and the obstacles are taken into account. 

Apart from static obstacles, there is a large number of papers dealing with collision avoidance against a moving obstacle using the APF method. Chen et al. \cite{Chen2013} present path planning based on the APF method for collision avoidance in a dynamic environment with faster response and high accuracy. The APF method applied in a dynamic environment is also presented in \cite{Sun2017}, \cite{Choi2020} and is out of the scope of this paper. Furthermore, the APF method can be used not only for the single robot path planning, but also for multi-robot systems, as described in \cite{Movric2010}, \cite{Oland2013}, \cite{Nascimento2014} and \cite{Zhang2019}. 

However, the main drawbacks of the APF method are oscillations, local minima and unreachable targets. To overcome the problem of local minima, the concept of the curl-free vector field is studied in \cite{Rezaee2012}, \cite{Choi2020}. In these papers, the curl-free vector field is utilized instead of the repulsive potential field, although the attractive potential field is the same as the conventional one. Additionally, in \cite{Choi2020}, the authors provide a solution to the problem of the unreachable target in the conventional APF method. 

Motivated by advances in literature and the fact that the APF method has been shown to perform well in obstacle avoidance but typically gets stuck in oscillations and local minima, this paper presents a novel algorithm based on potential fields to enable efficient path planning in a 3D environment.
Inspired by the idea in \cite{Choi2020}, we modified the APF method to resolve oscillations and local minima and focus on obstacle avoidance in a static environment with a single UAV. Similarly to the approach in \cite{Chiang2015}, we use a global trajectory planning method to find a path from the UAV position to the target and then activate the APF method for local path planning.  

\section{Proposed approach}
\label{sec:proposed}

\subsection{Problem Formulation}
The main goal of our approach is to autonomously and safely plan a path in a 3D environment using a UAV equipped with a sensory set that generates large point clouds, such as LiDARs. 
LiDARs provide point clouds in a large Field of View (FOV), which is crucial in obstacle avoidance. Autonomous navigation is performed using a UAV that has uncertain prior knowledge of the environment. This means that there is a possibility that the UAV will encounter obstacles on its flight trajectory that are unknown at the time of initial trajectory planning. In this work, we only consider static unknown obstacles in the environment. An example of such a scenario is conducting a flight mission in an earthquake zone. The initial trajectory planning is based on the old data of the environment, where the UAV may encounter previously unknown obstacles created by the debris of the earthquake. The main objective of the UAV is to reach the final goal point while secondary objective is to follow the originally planned flight path. Deviations from the originally planned flight path are allowed to avoid previously unknown obstacles in the area.

\begin{figure*}[t!]
	\centering
	\includegraphics[width=0.8\textwidth]{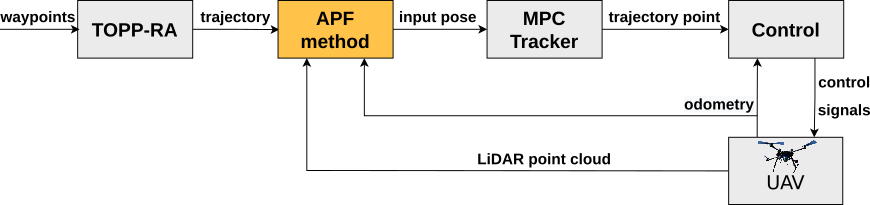}
	\caption{Overall schematic diagram of the path planning and obstacle avoidance system. The trajectory generated by TOPP-RA,\\ the LiDAR point cloud, and the odometry data represent inputs to the APF method. The MPC tracker module generates a trajectory\\ point to which the UAV navigates.}
	\label{fig:system}
\end{figure*}

\subsection{UAV and Sensor Models}
The UAV is represented with a state vector $\mathbf{x} = \begin{bmatrix} \mathbf{q}^T & \psi \end{bmatrix}^T \in \mathbb{R}^4$ that consists of the position $\mathbf{q} = \begin{bmatrix} x & y & z \end{bmatrix}^T \in \mathbb{R}^{3}$ and the yaw rotation angle around z axis $\psi \in [-\pi, \pi)$. Furthermore, the algorithm assumes a maximum linear velocity $\mathbf{v}_{max} \in \mathbb{R}^{3}$, a maximum angular velocity around z axis $\dot{\psi}_{max}$, a maximum linear acceleration $\mathbf{a}_{max} \in \mathbb{R}^{3}$ and maximum angular acceleration around z axis $\ddot{\psi}_{max}$.
The algorithm relies on a maximum range of the sensor $R_{max} \in \mathbb{R}$ with horizontal and vertical FOV in range, $\alpha_{h}$, $\alpha_{v}$ $\in (0^{\circ}, 360^{\circ}]$, respectively. This allows our algorithm to work with point-cloud-producing sensors with various FOV, such as camera with limited FOV and LiDARs with limited  $\alpha_{v}$.

\subsection{Global Trajectory Following}

The proposed system consists of a global and a local module. The global part of the system generates the trajectory by passing waypoints to the trajectory planner, while the local part utilize the APF method for the planned trajectory corrections in order to avoid obstacles. Within this paper, we use the Time Optimal Path Parametrization by Reachability Analysis (TOPP-RA) algorithm for planning a trajectory, which is developed in \cite{toppra}. Apart from the waypoints, inputs for the TOPP-RA are also velocity and acceleration constraints, which are maximally set to the UAV physical limitations. 

The trajectory is defined, among other parameters, by the number of trajectory segments, the time $t_k$, and the state $\mathbf{x}_k = \mathbf{x}(t_k)$ at knot point $k$.
The UAV simultaneously executes the generated trajectory and processes the data received from the sensor system. 
The proposed potential field algorithm calculates the total repulsive force $F_t$, which is described in detail in the next section.
Let us define $F_{threshold}$ as a constant that regulates whether the algorithm follows the trajectory point or activates the potential field algorithm. The condition is described as follows:

\vspace{0.1cm}
\text{Action} = 
$\begin{cases}
    \text{Follow Trajectory} & \text{if $ F_t < F_{threshold}$}, \\
    \text{Do APF Method} & \text{if $F_t \geq F_{threshold}$}.\\
\end{cases}$
\vspace{0.1cm}

Let the APF method trigger at the time $t_k$. 
While the APF method is active, the algorithm keeps track of the time needed for the APF method to be deactivated, e.g., to avoid the obstacle, denoted as $t_o$.
When the total repulsive force acting on the UAV falls below the predefined threshold $F_{threshold}$, the obstacle is considered to be bypassed. The UAV continues to fly in the direction of the originally planned trajectory, following the trajectory point with state $\mathbf{x}(t_k + t_o)$.

Note that if the UAV is progressing the APF method for a long time $t_o$ (e.g., when avoiding large obstacles), it may not be able to return to and continue following the optimal and originally planned trajectory. In other words, the trajectory point with the state $\mathbf{x}(t_k + t_o)$ is the last or almost last point of the trajectory. Nevertheless, in this paper we focus on the obstacle avoidance algorithm and leave the optimal trajectory execution to future work. 

An overview of the proposed system is given in Fig. \ref{fig:system}. When the APF method is not active, the global trajectory points are passed directly to the MPC tracker module. On the other hand, if the APF method is active, the modified path is generated and forwarded to the MPC tracker module.
A standard PID cascade is used to control the UAV, with the inner loop controlling the velocity and the outer loop controlling the position. In our case, the reference for the controller is a trajectory point.

\subsection{Local Path Planning using a Potential Field-Based Algorithm}

Local path planning integrates the potential field algorithm to correct the planned global trajectory and generate a safe and collision-free path. 

The conventional APF is composed of two types of potential fields: attractive potential field and repulsive potential field \cite{Khatib1986}. Attractive potential field $U_a(\cdot)$ is usually formed by goal location $\mathbf{q}_g$ so that the goal point attracts the UAV in the field. In contrast, repulsive potential field  $U_r(\cdot)$ is generated by obstacles’ positions $\mathbf{q}_o$ so that obstacles repel the UAV when the UAV moves close to a certain range ($d_0$) around obstacles. The sum of these two potential fields results in total potential field $U_t(\cdot)$ that directs the UAV toward the goal point while avoiding obstacles:

\begin{equation}
    U_t(\mathbf{q}) = U_a(\mathbf{q}) + U_r(\mathbf{q}).
    \label{eq: pot_field_original}
\end{equation}

In this paper, we propose a modified potential field algorithm that uses only the repulsive force. Namely, the attractive force is set to zero,  $U_a(\mathbf{q}) = 0$. The UAV follows the planned trajectory and the repulsive force corrects it when it collides with obstacles.

The $U_r$ is given by \cite{Khatib1986}:

\begin{equation}
  U_r(\mathbf{q}) = 
  \begin{cases}
        \frac{1}{2} k_{rt} (\frac{1}{d(\mathbf{q}, \mathbf{q}_o)} - \frac{1}{d_0})^2 & \text{if $ d(\mathbf{q}, \mathbf{q}_o) \leq d_o$}, \\
        0 & \text{if $ d(\mathbf{q}, \mathbf{q}_o) > d_o$}, \\
    \end{cases} 
\end{equation}
where $k_r$ is repulsive gain coefficient, $d(\mathbf{q}, \mathbf{q}_o)$ is the relative distance between the position of the UAV $\mathbf{q}$ and the obstacle $\mathbf{q}_o$ and $d_0$ is the limiting distance of the potential field influence.
The corresponding \textit{translational repulsive potential field force}, is derived by computing the negative gradient of the repulsive potential function as follows:

\begin{equation}
\begin{split}
& \mathbf{F}_{rt}(\mathbf{q}) = - \nabla U_r(\mathbf{q}) = \\
& \begin{cases}
         k_{rt} (\frac{1}{d(\mathbf{q}, \mathbf{q}_o)} - \frac{1}{d_0})\frac{1}{d(\mathbf{q}, \mathbf{q}_o)^3}(\mathbf{q} - \mathbf{q}_o) & \text{if $ d(\mathbf{q}, \mathbf{q}_o) \leq d_o$}, \\
        0 & \text{if $ d(\mathbf{q}, \mathbf{q}_o) > d_o$},\\
    \end{cases}
\end{split}
\label{eq:translation_force}
\end{equation}
where $k_{rt}$ is the gain of the translational force vector.
Since the repulsive potential field is produced for each obstacle, the total potential field for $n$ obstacles is expressed as the sum of the repulsive forces for $n$ obstacles:

\begin{equation}
    \mathbf{F}_t = \sum_1^n{\mathbf{{F}}_{rt_i}}.
    \label{eq: pot_field_total}
\end{equation}







Although the conventional APF method generates an effective path, it encounters several problems, such as oscillations in the presence of obstacles or in narrow passages and getting stuck in local minima \cite{Choi2020}.
Among the above drawbacks, the problem of local minima is considered in this paper as it leads to an incomplete path. 
Since we propose a modified APF method where the attractive force is always zero, $U_a(\mathbf{q}) = 0$, the total force acting on the UAV is never zero when the UAV is under the influence of the potential field. Problematic in this approach is the case where the direction of the UAV's velocity vector $\dot{\mathbf{q}}$ is opposite to the direction of the translational force vector $\mathbf{F}_{rt}$. As shown in the first simulation scenario, the UAV repeatedly enters and exits the region where the potential field is generated along the same axis, resulting in an oscillatory motion of the UAV around the local minima (Fig. \ref{fig:sim_wall} (top)).
\begin{figure}[t!]
	\centering
	\includegraphics[width=1\columnwidth, trim={1cm 1cm 1cm 2.5cm}, clip]{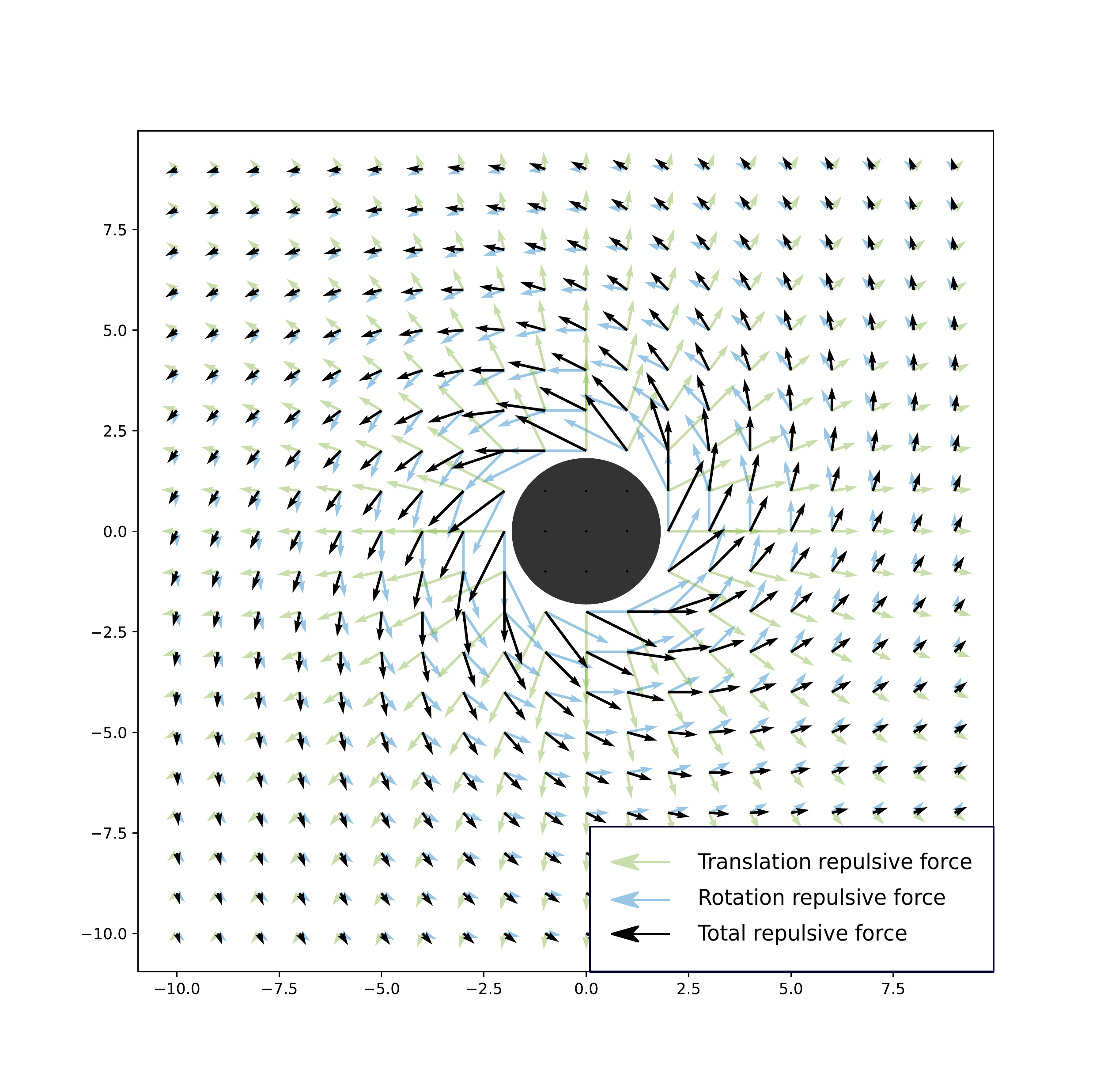}
	\caption{Repulsive force generated by the obstacle. The total repulsive force consists of the translational component vector $\textbf{F}_{rt}$ and the rotational component vector $\textbf{F}_{rr}$, where the rotational potential field is generated in the counterclockwise direction.}
	\label{fig:rep_force}
\end{figure}

\begin{figure}[t!]
	\centering
	\includegraphics[width=1\columnwidth, trim={0cm 1cm 0cm 2cm}, clip]{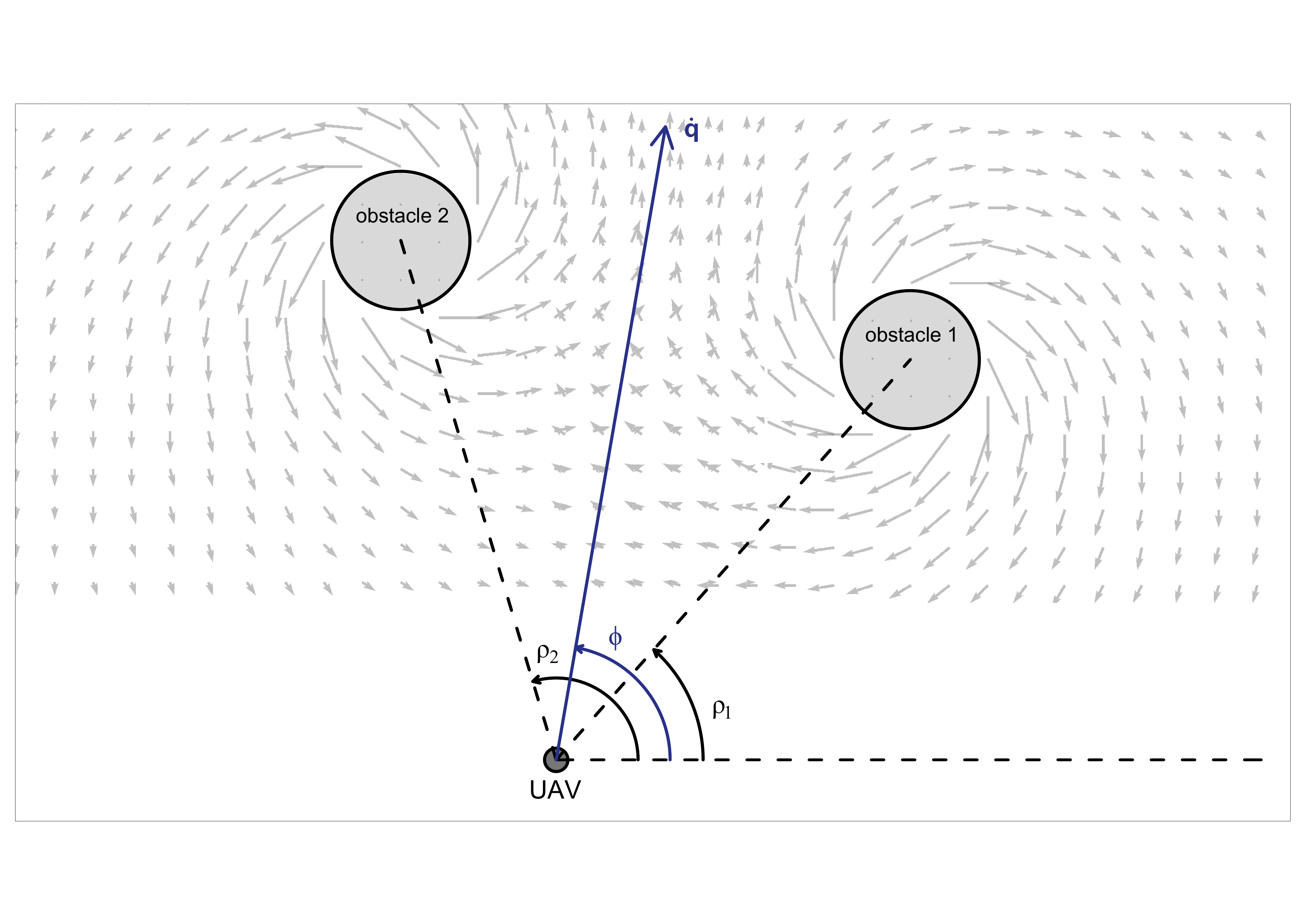}
	\caption{Direction of the rotational potential field generated by the obstacle $i$ in the environment is defined by the difference between the angle of the trajectory $\phi$ and the angle $\rho_i$, which is an angle of a vector from the UAV to the centroid of the obstacle $i$.}
	\label{fig:angles_cw_ccw}
\end{figure}

To overcome this drawback, we propose an extension of the repulsive force $\mathbf{F}_r$ to include both the translational and rotational components of the repulsive force.
The modified repulsive force is formulated as follows:

\begin{equation} 
    \mathbf{F}_r(\mathbf{q}) = \mathbf{F}_{rt}(\mathbf{q}) + \mathbf{F}_{rr}(\mathbf{q}),
\end{equation}

where $\mathbf{F}_{rt}(\mathbf{q})$ is the translational part of the repulsive force defined with Eq. \ref{eq:translation_force}, while $\mathbf{F}_{rr}(\mathbf{q})$ is the \textit{rotational repulsive potential field force} shown in Fig. \ref{fig:rep_force}. The rotational component of the repulsive force is calculated only in the X-Y plane, so we can write position vector of the UAV $\mathbf{q} = \begin{bmatrix} \mathbf{r} & z \end{bmatrix}^T \in \mathbb{R}^3$, position vector of the obstacle $\mathbf{q}_o = \begin{bmatrix} \mathbf{r}_o & z_o \end{bmatrix}^T \in \mathbb{R}^3$, and defined so that the curl of the potential field $U_{rr}(\mathbf{q})$ equals zero:
\begin{equation}
    \nabla \times U_{rr}(\mathbf{q}) = 0
\end{equation}
\begin{equation}
    \begin{split}
    & \mathbf{F}_{rr}(\mathbf{q}) = \\
    & \begin{cases} k_{rr} (\frac{1}{d(\mathbf{q}, \mathbf{q}_o)} - \frac{1}{d_0})\frac{1}{d(\mathbf{q}, \mathbf{q}_o)^3}R(\mathbf{r} - \mathbf{r}_o) & \text{if $ d(\mathbf{q}, \mathbf{q}_o) \leq d_o$}, \\ 0 & \text{if $ d(\mathbf{q}, \mathbf{q}_o) > d_0$},\\
    \end{cases}
    \end{split}
\end{equation}
where $k_{rr}$ is the gain of the rotational force vector and $R$ is the rotation matrix. The definition of the matrix $R$ depends on the direction of the generated rotating repulsive field, which can be clockwise or counterclockwise. The angle $\phi$ is defined as the angle of the UAV on the trajectory and $\rho$ is the angle of the vector from the position of the UAV to the centroid $C_i$ of the obstacle in the environment (Fig. \ref{fig:angles_cw_ccw}). The angle $\theta$ is defined as the difference between these two angles $\theta = \phi - \rho$ and its sign determines the direction of the potential field around the obstacle. The definition of the matrix $R$ with respect to the $\theta$ is given by \cite{Choi2020}:
\begin{equation} 
    R = 
    \begin{cases} 
        \left[\begin{array}{cc} 0 & 1 \\ -1 & 0 \end{array}\right] & \text{if $\theta \geq 0$}, \\
        \\
        \left[\begin{array}{cc} 0 & -1 \\ 1 & 0 \end{array}\right] & \text{if $\theta < 0$}. \\
    \end{cases}
\end{equation}



To find the centroid $C_i$ of each obstacle, we use a clustering method and the computation of a 3D centroid from the PCL library \cite{Rusu_ICRA2011_PCL}. A clustering method is used to divide a disorganized point cloud model of the environment into smaller parts that represent obstacles. The simplest way is to use the Euclidean clustering algorithm,  which is a greedy growing region algorithm based on the nearest neighbor. The cluster affinity is based on the distance to each point of a cluster and is defined by the parameter called cluster tolerance $c_{tolerance}$. If $c_{tolerance}$ is a very small value, an obstacle can be seen as several clusters. On the other hand, if the set value is very high, multiple obstacles can be seen as one cluster. Thus, the value of $c_{tolerance}$ is determined experimentally.
The 3D centroid $C_i$ is then computed for each cluster of the given point cloud. 

The proposed combination of forces ensures that the UAV always moves away from the obstacle and in the direction of the planned trajectory. This prevents the UAV from getting stuck in an oscillatory motion around the local minimum created by the translational forces or from orbiting the obstacle at the same distance as caused by the rotational forces. When the total repulsive force acting on the UAV falls below the predefined threshold $F_{threshold}$, the obstacle is considered to be bypassed and the UAV continues flying in the direction of the originally planned trajectory. 


\subsection{MPC Tracker}
An MPC-based tracking method is chosen to generate UAV trajectory points along the corrected collision-free path obtained from the potential fields. The original implementation is presented in \cite{tomas_mpc} while an adapted version of their work is used in this paper.

This tracking method employs a model predictive controller with a constant snap UAV model which controls a virtual UAV using snap commands. Snap commands are used as the input to the linear system, which predicts the next virtual UAV state based on the given model dynamics. The complete state of the virtual UAV is then sampled and used as a referent trajectory point for the real UAV at a rate of 100 Hz.

The definition of the MPC problem is as follows:
\begin{equation} \label{eq:mpc_qp}
\begin{aligned}
&  \underset{\textbf{u}_0, \, \dots \, , \textbf{u}_{\text{N}}}{\text{min}} & & 
\sum\limits_{\text{i} = 0}^{\text{N}}
    \left( 
        \textbf{e}_{\text{i}}^\text{T} \text{Q} \textbf{e}_{\text{i}} + 
        \textbf{u}_{\text{i}}^\text{T} \text{P} \textbf{u}_{\text{i}}
    \right)\\
& \text{s.t.} & &  \textbf{x}_{\text{k} + 1} =
\text{A} \textbf{x}_\text{k} + 
\text{B} \textbf{u}_\text{k} \, , \\
& & & \textbf{x}_\text{k} \leq \textbf{x}_{max} \, , \\
& & & \textbf{u}_\text{k} \leq \textbf{u}_{max} \, ,
\end{aligned}
\end{equation}
where $N$ is the horizon length. The error between the predicted virtual UAV state and the reference at the $k$-th horizon is defined as $\textbf{e}_\text{k}=\textbf{x}_\text{k} - \textbf{r}_{\text{k}}$. The state and input constraints are denoted by $\textbf{x}_{max}$ and $\textbf{u}_{max}$, respectively. The matrices A and B represent the well-known constant snap virtual UAV model. The weights Q and P are tuned for smooth trajectory generation wrt the velocity and acceleration.

The CVXGEN solver is used to obtain the optimal snap input $\textbf{u}^*_0$, which is used as an input to predict the next virtual UAV state and calculate the next referent trajectory point.

The main motivation for using this tracking method is that it allows us to quickly change and re-plan the UAV trajectory based on the current system state and model dynamics. Furthermore, the tracker enables safe and stable flight, regardless of the goal point resulting from the potential fields.
\section{Simulation-based evaluation}
\label{sec:simulation}

Simulations are performed in the Gazebo environment using Robot Operating System (ROS) and a model of the \textit{Kopterworx} quadcopter, which is identical to the one used for experiments in the real world. The quadcopter is equipped with a Velodyne VLP-16 LiDAR sensor. We run four scenarios with different complexity and analyze the results. 
Parameters used in the experiments are shown in Table \ref{tab:parameters}.
\bgroup
\def\arraystretch{0.6}
\begin{table}[h!]
\centering
\caption{The APF method parameters}
\label{tab:parameters}
\begin{tabular}{l *4c}
\toprule
Parameter & Scenario 1 & Scenario 2 & Scenario 3 & Scenario 4\\
\midrule
$k_{rt}$ & 153.0 & 153.0 & 95.0 & 800.0 \\
\midrule
$k_{rr}$ & 1720.0 & 1720.0 & 600.0 & 0.5 \\
\midrule
$\mathbf{v}^{\left\{x,y,z \right\}}_{max}$ [m/s] & 2.0 & 2.0 & 1.5 & 1.0 \\
\midrule
$d_0$ [m] & 15 & 15 & 15 & 10 \\
\midrule
$c_{tolerance}$ & 1.0 & 1.0 & 1.0 & 1.0 \\
\midrule
$F_{threshold}$ & 0.2 & 0.2 & 1.6 & 1.0 \\
\bottomrule
\end{tabular}
\end{table}
\egroup
\begin{figure}[t!]
	\centering
	\includegraphics[width=0.8\columnwidth, trim={3cm 0cm 3.5cm 0cm}, clip]{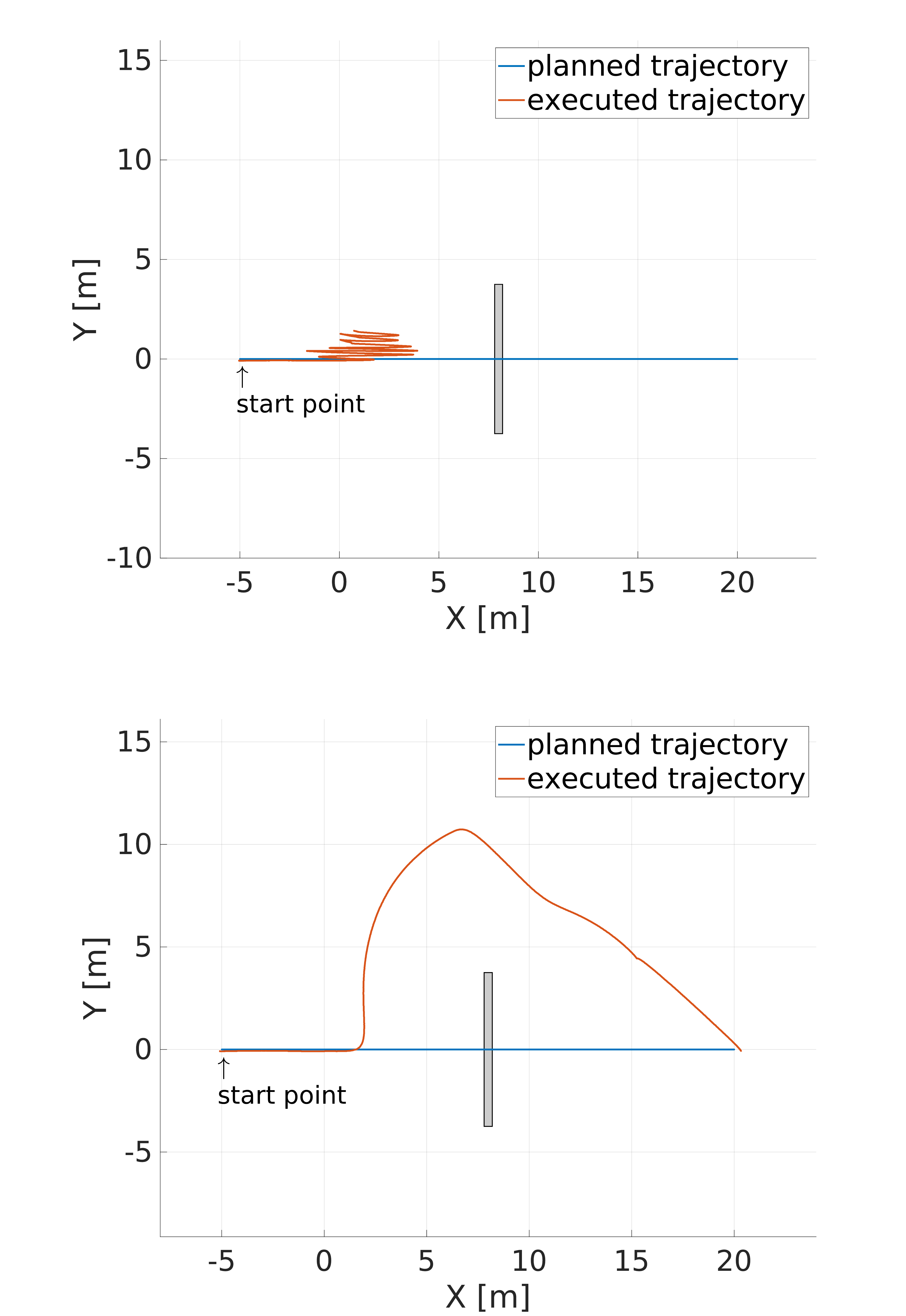}
	\caption{Simulation results in the X-Y plane for the case where the UAV trajectory is perpendicular to a large obstacle. In the case where the conventional APF was used for obstacle avoidance, the UAV got stuck at the local minimum (top). When the modified repulsion force with rotational component was used, the UAV generated a trajectory around the obstacle and successfully reached the target point (bottom).}
	\label{fig:sim_wall}
\end{figure}
\begin{figure}[t!]
	\centering
	\includegraphics[width=1\columnwidth]{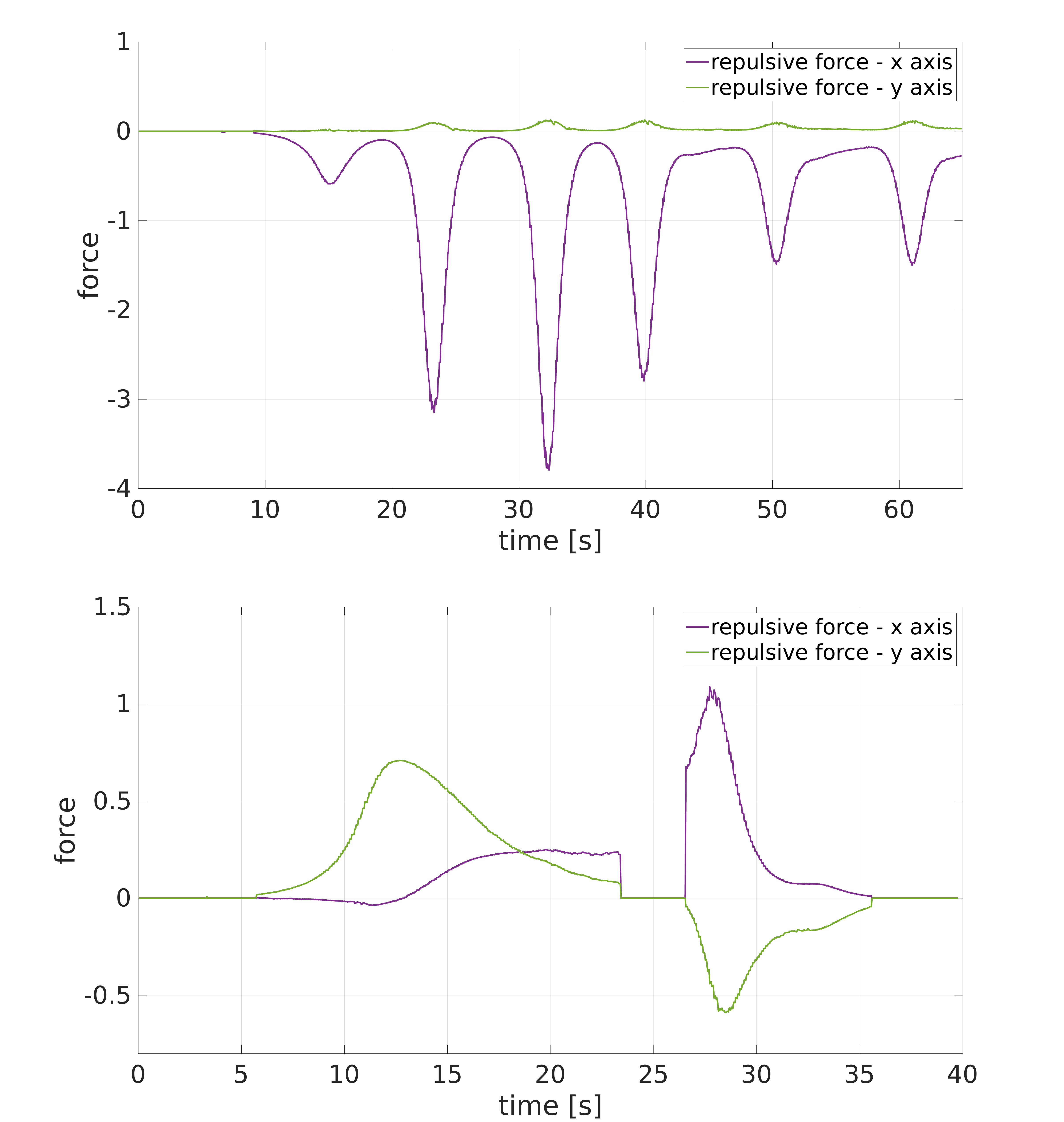}
	\caption{The total repulsive forces generated by an obstacle in the x and y axes during mission execution. The UAV gets stuck in local minima and performs oscillatory motion in the x direction (top), while the modified repulsive force with rotational component ensures successful mission execution (bottom).}
	\label{fig:sim_wall_force}
\end{figure}

The first scenario refers to a relatively simple space with a rectangular obstacle. As shown in Fig. \ref{fig:sim_wall}, the conventional APF method proposed by \cite{Khatib1986}, which includes only a translational repulsive potential field force, cannot guide the UAV to the target point. The UAV gets stuck in local minima and performs an oscillatory motion in the x-direction around local minima, as shown in Fig. \ref{fig:sim_wall_force}. The proposed APF method generates the corrected collision-free path around the obstacle (Fig. \ref{fig:sim_wall} (bottom)) and the MPC tracker ensures smooth and fast path execution. It can be observed that the gain coefficients $k_{rt}$ and $k_{rr}$ are set such that the total potential field force ensures that the UAV flies around the obstacle with a distance of nearly 5 m. In this case, the UAV does not manage to return to the originally planned trajectory because the size of the obstacle makes the detour longer, i.e., the value of the repulsive force  stays above the threshold $F_{threshold}$ longer.
\begin{figure}[t!]
	\centering
	\includegraphics[width=1\columnwidth]{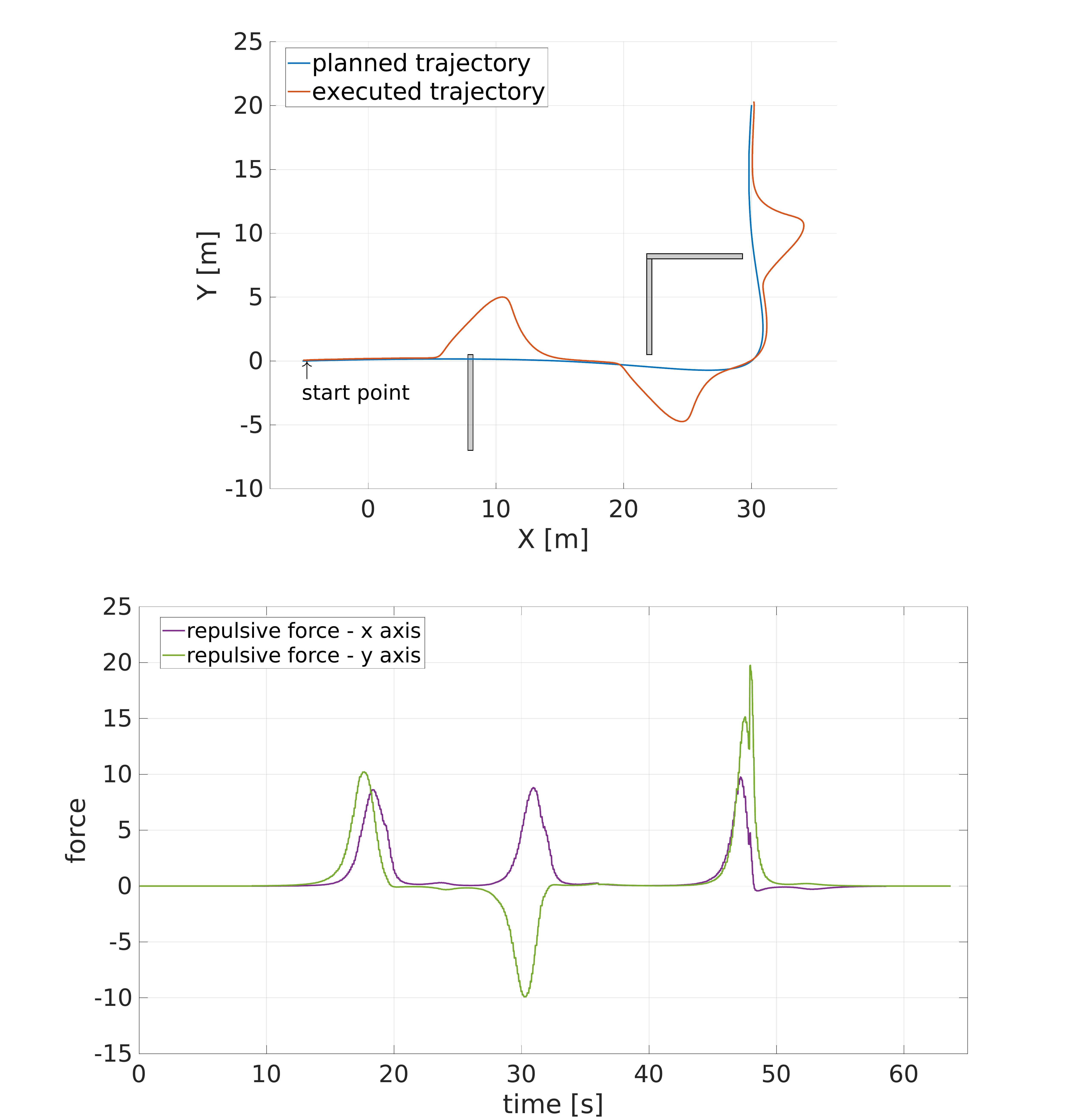}
	\caption{Simulation results in the X-Y plane (top) in an environment where only small disturbances occur during the execution of the flight trajectory. Since only small path corrections are required to avoid obstacles, the value of the repulsive force quickly drops below the threshold $F_{threshold}$ and the UAV manages to return to the originally planned trajectory before reaching final goal. }
	\label{fig:sim_wall_small}
\end{figure}

In the second scenario, only small flight path corrections are required to avoid obstacles in the environment (Fig. \ref{fig:sim_wall_small}). This means that the UAV is only under the influence of the force generated by the APF for a short period of time and is able to return to the originally planned trajectory before reaching the final goal point.

\begin{figure}[t!]
	\centering
	\includegraphics[width=1\columnwidth]{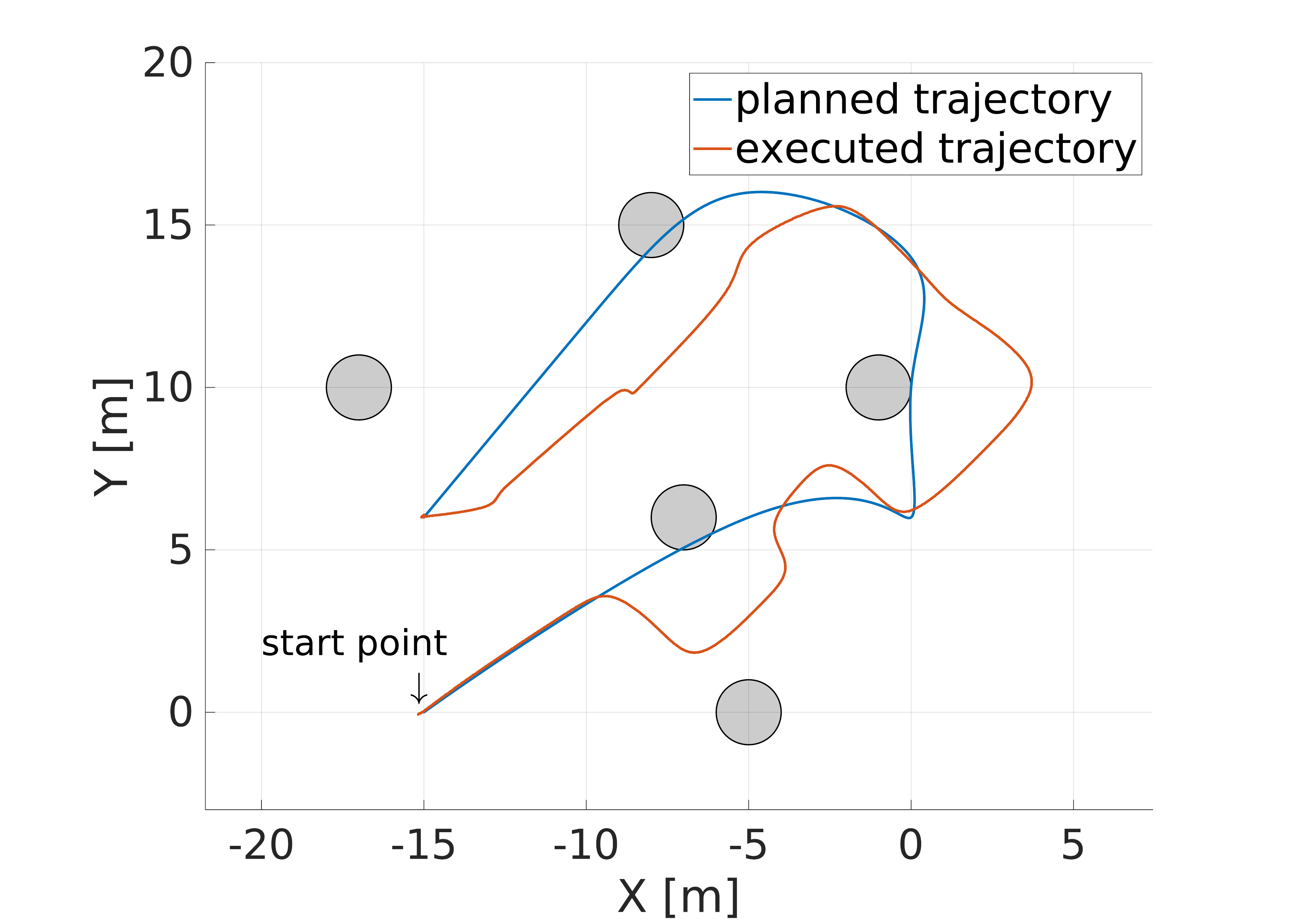}
	\caption{Simulation results in the X-Y plane in a complex environment with multiple obstacles near the planned flight path. During the execution of the flight, deviations from the planned trajectory occur that are so large that the UAV cannot return to the originally planned trajectory, but is still able to reach final goal point.}
	\label{fig:sim_cylinders}

\end{figure}

\begin{figure}[t!]
	\centering
	\includegraphics[width=1\columnwidth]{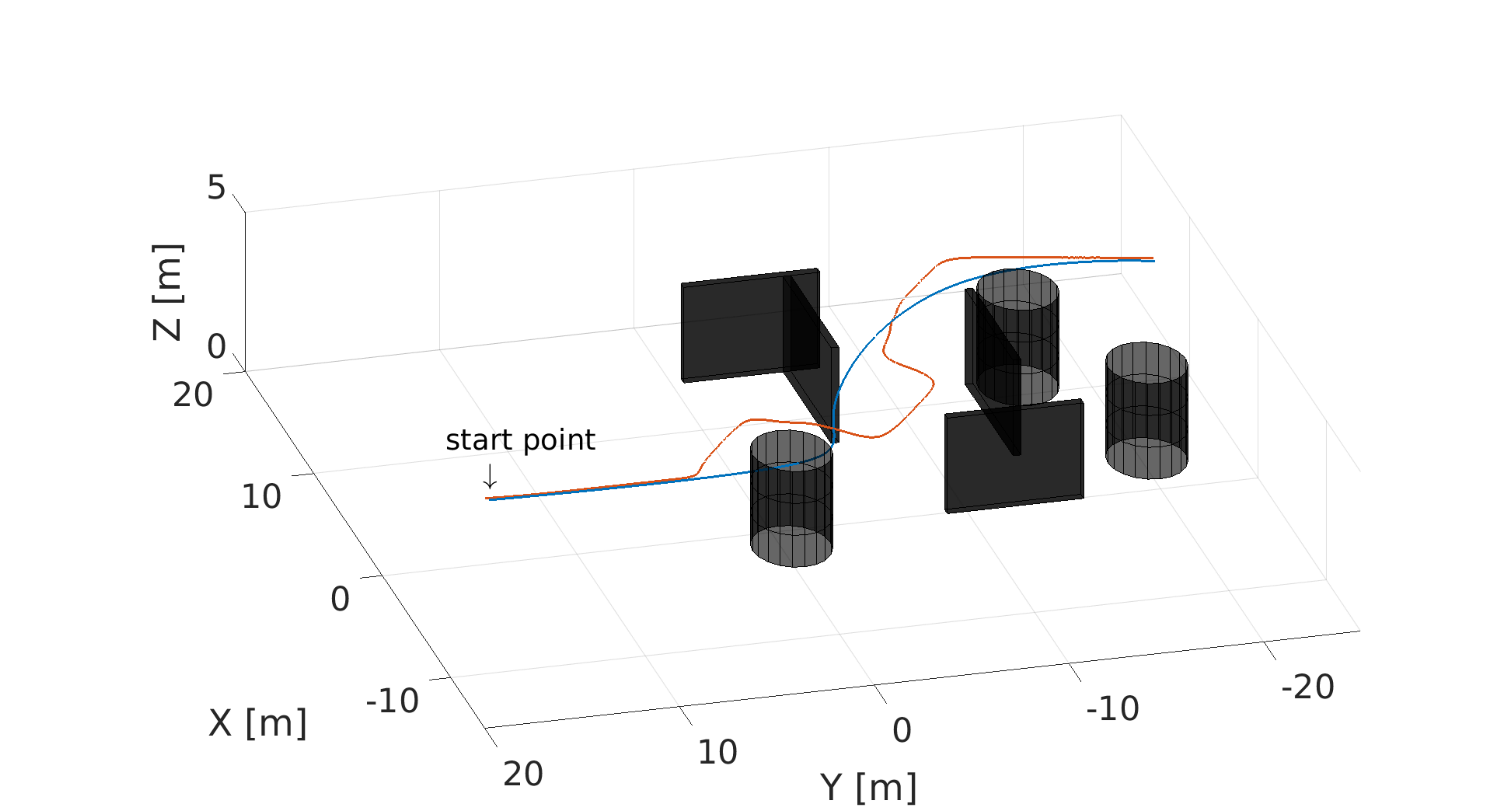}
	\caption{Simulation results in the fourth scenario in the maze-like environment with the different types of obstacles.}
	\label{fig:sim_lybirinth}
\end{figure}

The third scenario consists of cylinders through which a trajectory is planned. Fig. \ref{fig:sim_cylinders} shows the planned and the executed trajectory. The fourth scenario (Fig. \ref{fig:sim_lybirinth}) refers to a complex space with a maze-like environment. In both complex scenarios, obstacles in the environment create large deviations from the originally planned trajectories, but the UAV is still able to reach the final goal point. Each scenario confirms that the proposed APF method plans safe and collision-free paths.

\section{Experimental evaluation}
\label{sec:experiment}

\begin{figure}[t!]
	\centering
	\includegraphics[width=0.95\columnwidth]{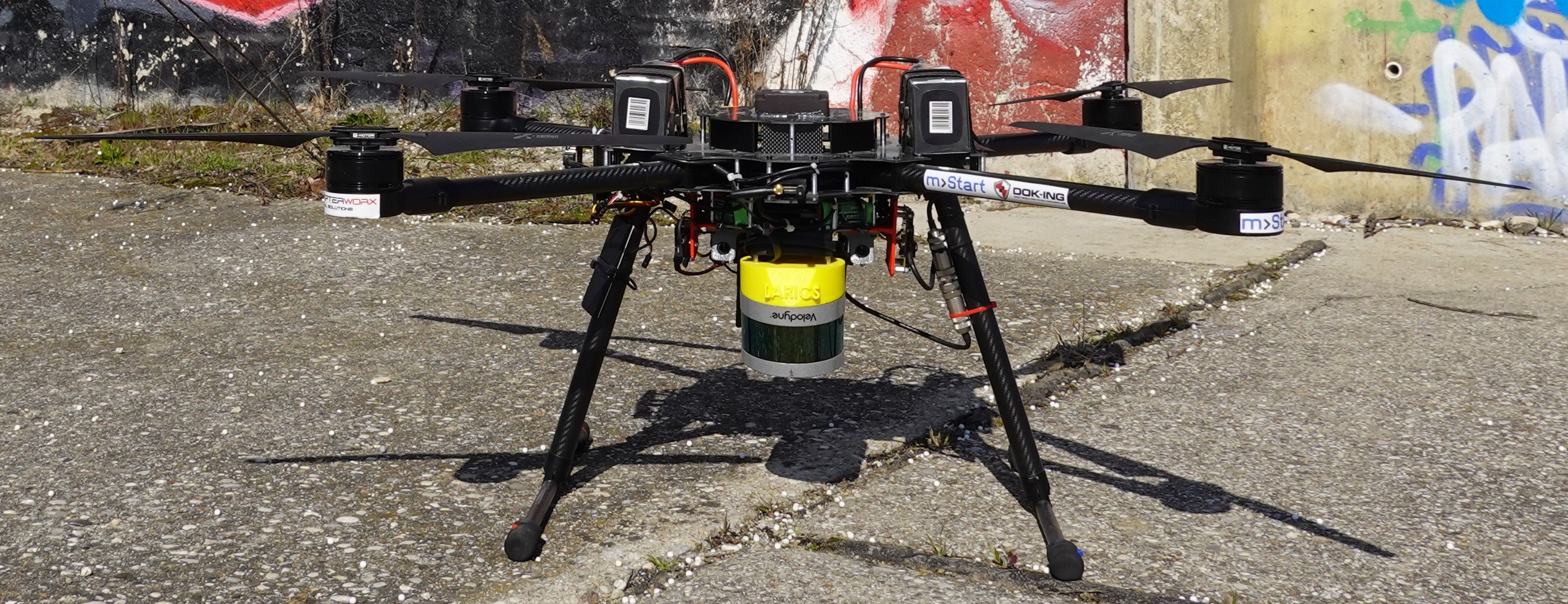}
	\caption{A custom built quadcopter equipped with a Velodyne $VLP-16$ LiDAR sensor.}
	\label{fig:uav}
\end{figure}

For our outdoor experimental analysis, we use a \textit{Kopterworx} quadcopter (Fig. \ref{fig:uav}) which features four \textit{T-motors} P60 KV170 motors attached to a carbon fiber frame. The dimensions of the UAV are 1.2 m $\times$ 1.2 m $\times$ 0.45 m, which makes it a relatively large UAV suitable for outdoor environments. The total flight time of the UAV is around 30 min with a mass of m = 9.5 kg, including batteries, electronics and sensory apparatus. The \textit{Pixhawk 2.1} flight controller unit is attached to the center of the UAV body, and it is responsible for the low-level attitude control of the vehicle. Furthermore, we equipped the UAV with an \textit{Intel NUC}, i7-8650U CPU @  1.90GHz $\times$ 8, on-board computer for collecting and processing sensory data. The on-board computer runs \textit{Linux Ubuntu 18.04} with \textit{ROS Melodic} framework that communicates with the autopilot through a serial interface. The UAV is equipped with a Velodyne $VLP-16$ LiDAR sensor with a maximum range of 100 m. The maximum velocity of the UAV is limited to 1.0 m/s with a maximum acceleration of 0.5 m/s$^2$.

\begin{figure}[t!]
	\centering
	\includegraphics[width=1\columnwidth]{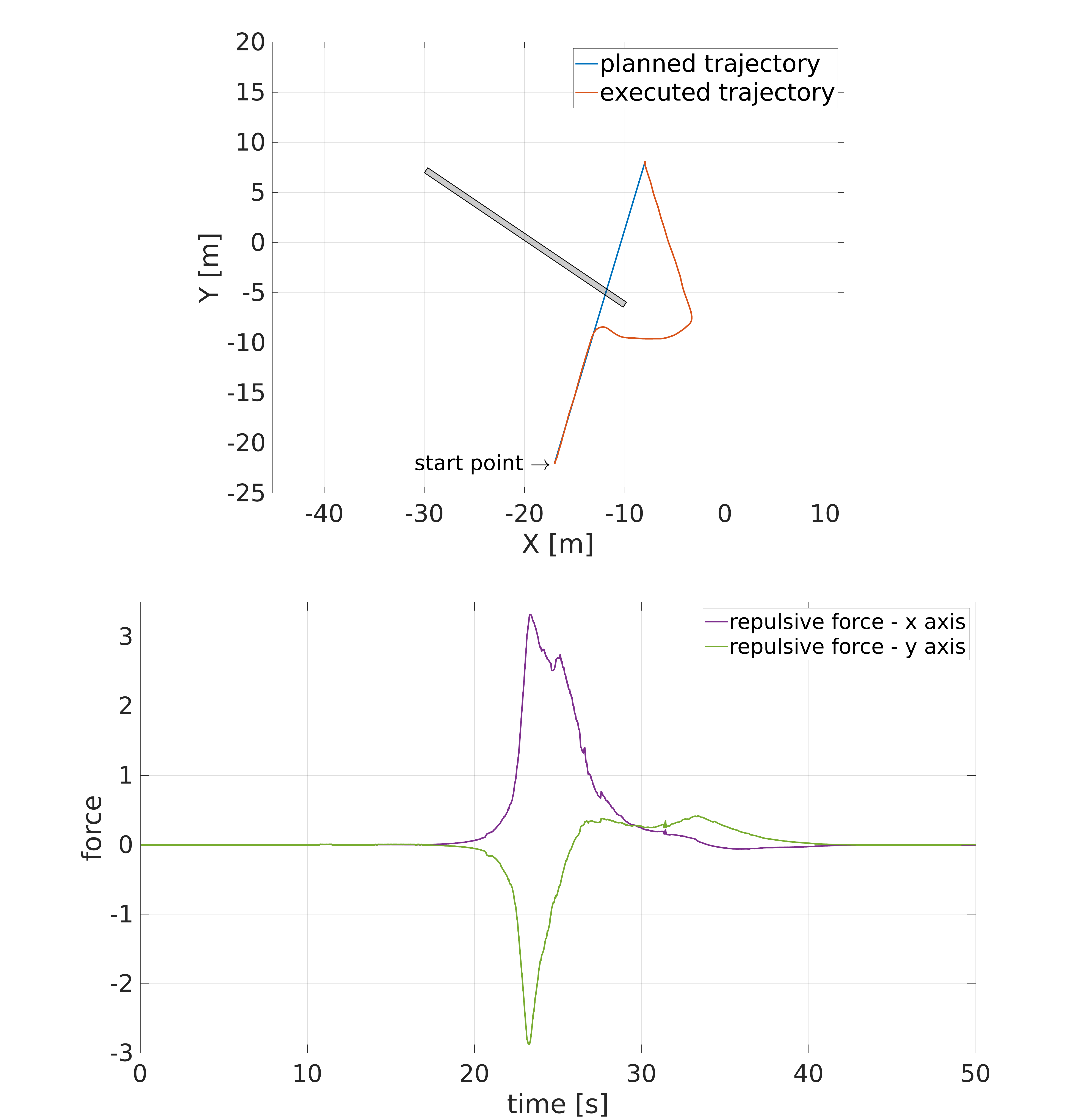}
	\caption{The results of the real-world experiments show that the UAV successfully avoids the wall on its flight trajectory by using the proposed method for modified APF with rotational and translational force.}
	\label{fig:experiment_results}
\end{figure}

Fig. \ref{fig:experiment_results} shows the results of the real-world experiment, which is similar to the first simulation scenario.

\section{Conclusion and future work}
\label{sec:conclusion}

This paper deals with a modified APF method for efficient and safe path planning. The method is capable of autonomous path planning and online obstacle avoidance. Furthermore, our modified algorithm uses a rotation-based component to avoid local minima.
The MPC tracker allows us to correct the UAV trajectory based on the current system state and the model dynamics.
The results show an improved behaviour in terms of trajectory execution and resolution of local minima compared to a state-of-the-art strategy. 
This 3D path planner has been successfully tested in simulation scenarios, as well as in a real-world experiment, using a quadcopter equipped with a LiDAR. 

In the future, we plan to test our method in a complex outdoor environment and extend our algorithm to avoid dynamic obstacles.

\appendices
\bibliographystyle{ieeetr}
\typeout{}
\balance
\bibliography{main}

\end{document}